\documentclass[a4paper]{article}

\usepackage{amssymb}
\usepackage{amsmath}
\usepackage{color}
\usepackage{multirow}
\usepackage{subfigure}
\usepackage{graphicx}
\usepackage{float}

\usepackage[
backend=biber,
style=authoryear,
citestyle=authoryear
]{biblatex}
\begin{document}

\title{NeuroClean: A Generalized Machine-Learning Approach to Neural Time-Series Conditioning}

%% Authors
\author{Manuel A. Hernandez Alonso$^{1}$, Michael Depass$^{1}$, Stephan Quessy$^{4}$,\\Ali Falaki$^{4}$, Soraya Rahimi$^{4}$, Numa Dancause$^{4}$, Ignasi Cos$^{1,2,3}$}

\clearpage

\maketitle

\vspace{1cm}

\begin{flushleft}$^{1}$ Departament de Matemàtiques i Informàtica, Universitat de Barcelona, Gran Via de les Corts Catalanes 585, 08007 Barcelona, Catalonia, Spain\\
$^{2}$ Institut de Neurociència UB, Barcelona, Catalonia, Spain\\
$^{3}$ Institut de Matemàtiques UB, Barcelona, Catalonia, Spain\\
$^{4}$ Department of Neuroscience, Chemin de la Tour, Montreal, QC H3T 1J4, Canada\\
* corresponding author: manuel.hernandez@ub.edu
\end{flushleft}

\vspace{1cm} 

\clearpage

\begin{abstract}
Electroencephalography (EEG) and local field potentials (LFP) are two widely used techniques to record  electrical activity from the brain. These signals are used in both the clinical and research domains for multiple applications.
However, most brain data recordings suffer from a myriad of artifacts and 
noise sources other than the brain itself. Thus, a major requirement for 
their use is proper and, given current volumes of data, a fully automatized 
conditioning. As a means to this end, here we introduce an unsupervised, multipurpose EEG/LFP preprocessing method, the NeuroClean pipeline. In addition to its completeness and reliability, NeuroClean is an unsupervised
series of algorithms intended to mitigate reproducibility issues
and biases caused by human intervention.
The pipeline is designed as a five-step process, including the common 
bandpass and line noise filtering, and bad channel rejection. However,
it incorporates an efficient independent component analysis with an automatic component rejection based on a clustering algorithm. This machine learning classifier is used to ensure that task-relevant information is preserved 
after each step of the cleaning process. We used several data sets to validate the pipeline. NeuroClean  removed several common types of artifacts from the signal. Moreover, in the context of motor tasks of varying complexity, 
it yielded more than 97\% accuracy (vs. a chance-level of 33.3\%) in an optimized Multinomial Logistic Regression model after cleaning the data, compared to the raw data, which performed at 74\% accuracy. These results show that NeuroClean is a promising pipeline and workflow that can be applied to future work and studies to achieve better generalization and performance on machine learning pipelines.
\end{abstract}

\clearpage

\section{Introduction}

Electroencephalography (EEG) and Local Field Potential (LFP) data are two
widely used techniques to record electrical activity from the brain. 
Either recorded by means of scalp electrodes in the case of EEGs or of
surgically implanted electrodes in specific brain areas, the outcome
are time-series of brain activity, along with several sources of noise
and multiple artifacts, typically recorded to investigate either normal
brain function or the contribution of specific areas to cognition or movement.
EEGs, furthermore LFP data, are highly dimensional data, as their signals
gather aggregate contributions from groups of neurons firing simultaneously 
with a high sampling rate via multiple channels. 
Consequently, physiological and non-physiological noise and artifacts are 
common occurrences when working with EEG and LFP data (\cite{Eeg-guidelines}). 
Typical noise sources are electrical wiring, eye blinks, muscle spasms, 
eye movements, which vary greatly in intensity and statistics. Consequently, 
systems and schemes have been developed to remove and mitigate these noisy 
components. However, most of them involve long
hours of semi-manual labor for a researcher to mitigate these problems.
This task requires domain-specific knowledge, with automatized and flexible 
pipelines that offer consistent, objective, and standardized data cleaning, 
with an efficient data workflow for large datasets. Nevertheless, automated 
pre-processing pipelines raise several hard challenges in finding the most 
reliable method to distinguish good from bad data. In particular, we may 
inadvertently remove neural data losing some fine details of brain activity, 
relevant to further analyses. Similarly, failing at removing noise and artifacts 
may result in noise biases when fitting models, and poor reproducibility of 
research investigations across experiments in different laboratories. 

Previous work has created various automatic pipeline schemes that remove 
specific artifacts, or sets of these artifacts. Most of them make use of 
Independent Component Analysis (ICA), or variations thereof (such as wavelet 
enhanced ICA, wICA; \cite{nolan2010}, \cite{happe}, ), to reject artifactual components. Significant effort has 
been devoted to design pipelines for artifact rejection and noise mitigation (\cite{happe}; \cite{prep}; \cite{apice}, \cite{adjust}). Furthermore, some of the pipelines 
require pre-trained labeled models or structures that may not be
available with the specifics of each recording setup arrangement
(spatial data, electrode placement, etc.).  

To minimize the amount of neural data removed, quantifiable signal 
quality assessment metrics are needed. 
Previous work has shown several metrics and performance values for 
the different pipelines (\cite{eegquality}; \cite{happe}; \cite{prep}), 
some focused on visual differences and final signals, and others based 
on hyperparameter values, i.e., Signal-to-Noise Ratio (SNR), 
probability of noisy components, number of channels rejected, etc. 
In addition to these, here we introduce the use of a practical 
machine learning set-up to assess the utility of the pipeline 
in live analysis scenarios by tracking the performance of different 
classification models. Furthermore, this kind of ML analysis provides
the flexibility of being capable of coping with different datasets
recorded from various parts of the brain and targets of interest
(language, motor, attention, etc.). 

The ultimate goal of the NeuroClean pipeline is to provide a general, standardized, 
and effective workflow, usable across different experimental setups, which can be 
tailored to EEG and LFP datasets within virtually any experimental context.
We also add a new way to understand the effects of neural data cleaning via classification and model performance across different tasks and subjects. In particular, the NeuroClean pipeline has been tested on high-dimensional data of 256 channel LFP data in motor activity experiments performed on two macaque monkeys. However, the workflow and pipeline scheme may be able to properly fit other neural analysis fields. NeuroClean is implemented from the ground up and is freely available in Python with multiple integrations of MEEG-Kit and scipy. Thus, NeuroClean is a mostly automatic pipeline that processes data in full or in batches until completion. 
Furthermore, NeuroClean requires time-series alone, making it more accessible to datasets with no spatial or additional information.  The next sections provide a description of 
the specific implementation algorithms and steps used in this pipeline, as well as the ML
performance and signal quality assessments prior to the pipeline and after each individual
step of NeuroClean. 

\begin{figure}[ht]
\begin{center}
\setlength{\unitlength}{1pt}
\footnotesize
\includegraphics[width=8cm]{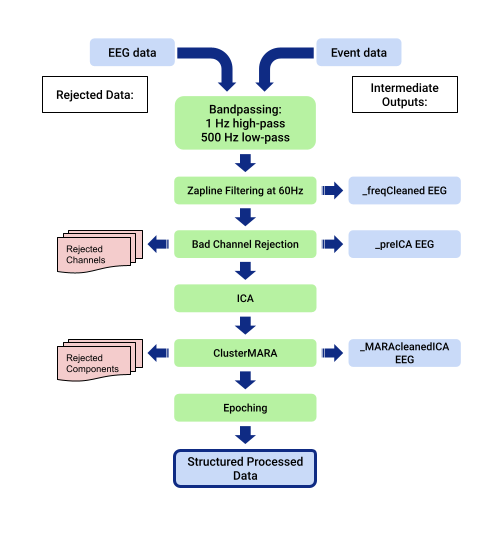}
\caption{\textit{Schematic of the preprocessing pipeline. It starts with a bandpass filter from 1Hz to 500Hz, followed by a zapline filter to remove power supply artifacts and their harmonics, then a bad channel rejection algorithm is applied, followed by an ICA with ClusterMARA to reject components, and finally the data is epoched to get a structured processed data.}}
\end{center}
\label{fig:high-level}
\end{figure}

\section{Materials and Methods}

\subsection{The NeuroClean Pipeline}

NeuroClean is a multipurpose neural data preprocessing pipeline based on algorithms that clean EEG/LFP signals. The pipeline input are multi-channel time series, and their respective meta-data, namely the sampling frequency, classes information (optional), and timed events (optional). The pipeline can output either a complete signal or a signal epoched into different classes. After running the NeuroClean pipeline, a file is generated with the desired output and a log is
provided. A schematic of the NeuroClean preprocessing pipeline is provided in \ref{fig:high-level}.

\paragraph{\textbf{Bandpass Filter}}

Bandpass filtering is applied in this pipeline to reduce noise around the signal and focus on brain wave data. That occurs mainly in relevant frequency bands around the alpha, beta, and low-gamma bands. Thus, a Butterworth filter is applied from 1Hz to 500Hz, following steps similar to \cite{daud2015butterworth}; where it was found that the application of this filter reduces some of the noise produced by different sources. For any signal that has a lower than 500 Hz sampling rate, the upper band of the band-pass filter is ignored. 

\paragraph{\textbf{Line noise filter}}

Then the ZapLine filter is used to remove the power supply noise from the EEG, as well as line noise that is found within the signal. This method was developed in \cite{de2020zapline} as a simple yet effective way to remove power line noise and its harmonics from the EEG signal. However, the ZapLine filter requires the frequency of the power supply (60Hz or 50Hz) to find the main frequency and its harmonics. A mix of spectral and spatial filtering is used when the filter is applied by means of two matrix decomposition branches where the target frequencies are removed. To be concise, it is assumed that the first branch $\mathbf{X}'$ is perfectly devoid of line artifacts, as the frequencies and harmonics of the line frequency are set to zero. On the other hand, the second branch $\mathbf{X}''$ is denoised by using a denoising matrix based on the JD/DSS method (\cite{de2014joint}). Lastly, the two branches are mixed back together by adding them. As shown in \cite{de2020zapline}, since the ZapLine method is based on a $1/f_{line}$ squared-shaped filter, the data not on the relevant denoised frequencies is barely affected. The filter was applied with the MEEGkit collection implemented for \textbf{Python 3.8+}.

\paragraph{\textbf{Bad Channel Rejection}}

With the preliminary filtering performed, it is necessary to reject broken, artifact-ridden, or otherwise unusable channels. We remove them by setting the corresponding channels to zero. Broken channels are found using standard deviation (SD) features and an iterative algorithm first described in \cite{KomosarFiedlerHaueisen}. Firstly, the algorithm starts by calculating the standard deviation of the signal for the $j$-th channel over the entire sample size of $N$, equation \ref{eq:sd_bcr}.

\begin{equation}
    \mathbf{S D}_{\mathbf{j}}=\sqrt{\frac{1}{N-1} \sum_{i=1}^{N}\left|V_{(i, j)}-\bar{V}_{j}\right|^{2}}
    \label{eq:sd_bcr}
\end{equation}

Where $V_i$ is the $i$-th sample of the $N$ samples for the $j$-th channel and $\bar{V}_{j}$ is the mean voltage for the $j$-th channel.

The algorithm then iteratively removes channels based on three different criteria: the median normalized SD is above the 75th percentile of the total population of SD, the standard deviation is below $<10^{-1}\mu V$, or the SD is above $>100\mu V$. An extra two criteria are used to end the iterations: the number of iterations of the algorithm exceeds 5, and if the number of bad channels detected in the last iteration is zero. 

\paragraph{\textbf{ICA and Cluster-MARA}}

This step is divided into two parts: first, an independent component analysis (FastICA in \cite{fastica}) is performed to extract the relevant sources of the data; then a modification of the Multiple Artifact Rejection Algorithm (MARA) proposed in \cite{maraalgo} is applied. To be precise, the modification of the MARA consists of using feature clustering instead of pre-trained models. This strategy combines the advantages of being a powerful supervised feature-based component rejection algorithm and a clustering method that can be fitted to any EEG cleaning case. In addition, previous work on ICA component rejection algorithms is based on supervised learning of manual classification data bases (\cite{happe}; \cite{winkler2015}; \cite{maraalgo}); this dependencies on other similar data may result on lower performance on unseen data, and in some cases (\cite{happe}) require additional information on the EEG recordings that may not be available.

Thus the Cluster-MARA algorithm is created, with a mix and selection of different features from the original MARA. These features are extracted from the component's time series, spectrum, pattern, and weight vector. Additionally, the original MARA had a spatial data related feature, which we discarded since it would require extra information other than the time series. Finally, the Cluster-MARA features are:

\begin{itemize}
    \item \textbf{Spatial range within pattern}, defined as the logarithm of the difference between the maximal and minimal activation of the weight vector.
    \item \textbf{Average log band power between 8 and 13 Hz}, defined as the average log band power of the alpha band between 8 Hz and 13 Hz. The power is computed using the Welch method to estimate the spectral power density of the signal.
    \item \textbf{Lambda $\lambda$}, describes the variance parameter of the fitting of the power spectral density of the component to the prototypical 1/frequency curve observed in brain waves.
    \item \textbf{Fit Error}, similar to $\lambda$, this feature describes the error in fitting the component's power spectral density to the prototypical 1/frequency curve.
    \item \textbf{Average local skewness in 15-second windows}, as the name suggests, a 15-second sliding window is applied to the time series of the component where the skewness is computed, then every computation is averaged to obtain the mean absolute local skewness of the component.
\end{itemize}

\begin{figure}[ht]
\begin{center}
\setlength{\unitlength}{1pt}
\footnotesize
\includegraphics[width=8cm]{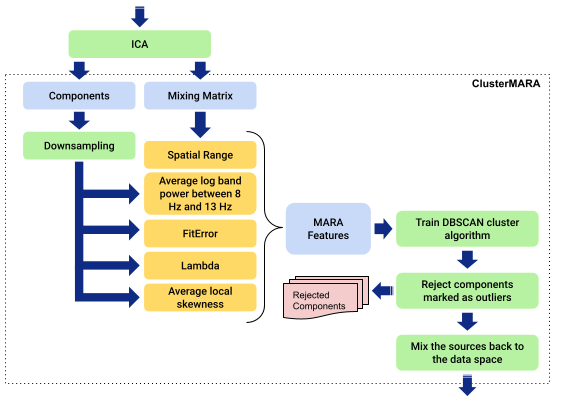}
\caption{\textit{A summary schematic of the ClusterMARA algorithm, a modification on the MARA algorithm presented by \cite{maraalgo}}}
\end{center}
\label{fig:marafeatures}
\end{figure}

After computing the features, we are left with a matrix of $s$ components by five features that is used to find clusters in this five-dimensional space. To achieve this, we apply Density-Based Spatial Clustering of Applications with Noise (DBSCAN, \cite{dbscan}) to the matrix, this finds the core samples of high density and expands clusters from them. In this DBSCAN method, a parameter representing the maximum intersample distance to be considered part of the same neighborhood was adjusted to a value of 2; another parameter related to the minimum number of samples in a neighborhood was also set to 2.

Then each component was marked and labeled with different cluster labels, or as outliers. The method used in this paper was to reject the outliers marked by the DBSCAN. However, there exists an interesting exercise and expansion possibility of iterating over the combination of clusters and rejecting each combination. This is due to the assumption that artifactual components will be only outliers, but it could be the case that there exist entire clusters of artifactual components. Finally, after rejecting the components by setting the component of the time series to zero, we mix the data back using the mixing matrix given by the FastICA algorithm. Finally, the implementation of this two-part step was performed using the FastICA function in ScikitLearn, and the extraction of MARA features was performed directly using SciPy and NumPy; additionally, DBSCAN was fitted through the implementation of ScikitLearn.

\paragraph{\textbf{Epoching (Optional)}}

As a last optional step, the pipeline can take classification information to segment the time series into trials and classes. The NeuroClean pipeline standardizes the length of the classes by centering the samples of each class by the given timestamps. Additionally, the pipeline discards any class that does not reach $p$ time-points, this parameter was tweaked to 500 with the data that was used to perform the tests.

\subsection{EEG Cleanliness Quality Assessment}

After executing the pipeline, a quality assessment on the input and output is performed in a per-step basis, as well as the general performance of the pipeline. We separate the quality assessment methods into two distinct sections: a traditional one with metrics such as Signal-to-Noise ratio, probability of artifacts, similarity to expected power curves, etc. (\cite{happe}; \cite{eegquality}); on the other hand, a machine learning pipeline was created to evaluate the empirical performance of the pipeline.

\subsubsection{Traditional Approach}

Traditional quality assessment methods were used to evaluate the effectiveness of each stage of the NeuroClean pipeline using well-established signal-based metrics. These methods provide quantitative measures of data cleanliness and are independent of downstream model performance, thereby offering a complementary validation framework to machine learning analysis.

First, the \textbf{Signal-to-Noise Ratio (SNR)} was calculated before and after each processing step to quantify the relative improvement in signal quality. The SNR was calculated as the ratio between the mean power of the cleaned signal and the residual noise component, following the method described in \cite{eegquality}. An increase in SNR across consecutive stages of the pipeline indicates effective attenuation of non-neural noise sources while preserving task-relevant neural oscillations.

Second, we evaluated the \textbf{artifact probability} for independent components identified during the ICA step. For each component, features such as local skewness, spectral slope, and deviation from the canonical $1/f$ spectral shape were used to estimate the likelihood of being artifactual. These probabilities were derived using a thresholding approach inspired by \cite{maraalgo}, which allows for an interpretable estimation of the proportion of components removed versus retained.

Third, we assessed the similarity of the cleaned power spectral density (PSD) to the expected \textbf{$1/f$ spectral distribution}, a key characteristic of physiological neural activity (\cite{eegquality}). This similarity was quantified using the Pearson correlation coefficient between the empirical PSD and the theoretical $1/f$ model, calculated for each channel and averaged across the dataset. A higher correlation value indicates better preservation of the physiological spectral structure after artifact removal.

Finally, two aggregate measures were included to provide a holistic view of the preprocessing efficiency: (1) the \textbf{percentage of channels retained} after bad channel rejection, serving as an indicator of data preservation, and (2) the \textbf{proportion of ICA components rejected} by Cluster-MARA, reflecting the degree of artifact contamination in the original data. These metrics, combined with SNR, artifact probability, and PSD similarity, produce a comprehensive description of the impact of each NeuroClean preprocessing stage on data integrity and quality.

\subsubsection{Performant Machine Learning}

A machine learning pipeline was used over the structured preprocessed data provided by the NeuroClean pipeline. This step was required in order to assess the cleanliness of the signal. The pipeline starts by segmenting the data into train and test splits of 80\% training and 20\% testing. Subsequently, the mean spectral amplitude $\overline{\mathbf{A}}$ is calculated for each channel of the data:

\begin{equation}
    \overline{\mathbf{A}}_{\mathbf{i}}=\frac{\sum_{j=1}^{N}\left|V_{i, j}\right|}{N}
\end{equation}

Where $\overline{\mathbf{A}}_i$ is the mean spectral amplitude for the $i$-th channel, $V_{i,j}$ is the voltage at the $j$-th sample in the $i$-th channel, and $N$ is total number of samples. With this, we end up with a vector matrix of features for each class sample. The previous step before processing is to balance the data so that we have $x$ number of class samples and $y$ labels corresponding to each class sample.

The data is then used to train a Multinomial Logistic Regression (MLR) model. This trained model then has trained coefficients of the features in the decision function. By ordering the trained coefficients in descending order, we get a variable that shows the importance of each feature, since the closer the feature is to zero, the less effect the feature has over the decision function. With this in mind, the coefficient indices are saved to use as indices of the most important characteristics.

Using this ordered indices list, we train four models iteratively, two 1-Nearest-Neighbors (1NN) model, and two MLR models. One of each kind of model is assigned to be trained with shuffled labels to determine if the results of the shuffled data coincide with the expected theoretical random model accuracy. Lastly, an increasing search is performed on the remaining models; they are trained with the first $d$ features of the data $X$ until we get to the total number of features $D$. In addition, we do a K-Fold cross-validation on each fitting of the models to reduce variance on the accuracies.

The goal of this step is to obtain the accuracies of each number $d$ of features to determine the optimal number of features. The iteration works as follows:

\begin{enumerate}
  \item Let $d = 0$.
  \item Let $d$ be the number of features for the current iteration.
  \item Let $X_d$ be the data with only the number of features $d$.
  \item Fit an MLR and a 1NN model with the $X_d$ data and the $y$ class labels.
  \item Let $a_{(d,r)}$ be the accuracy of the models with respect to the number of features $d$ and repetition $r$, calculated by predicting the class labels for the training data $X_{t,d}$ with the $d$ features.
  \item Let $y_s$ be a random permutation of $y$.
  \item Fit an MLR and a 1NN model with the $X_d$ data and the $s_y$ class labels.
  \item Let $a_{s(d,r)}$ be the accuracy of the shuffled models with the number of features $d$ and repetition $r$.
  \item If the repetition $r$ is less than the number of repetitions $R$, return to 4.
  \item Let $\overline{a}_d$ and $\overline{a}_s(d)$ be the mean accuracies of the models with the number of features $d$, calculated by predicting the class labels for the training data $X_{t,d}$ with the $d$ features.
  \item If the number of features $d$ is less than the number of features $D$, then increment $d$ and return to 2.
\end{enumerate}

Additionally, we implemented a tolerance and epsilon method that will stop the iteration early if the improvement in the last 30 features has not improved the accuracy by at least an $\epsilon$ number, i.e. $10^-3$.

\section{Results}

\subsection{Experiment Setup}

The \textbf{first dataset} used was a reach-to-grasp behavioral task first described in \cite{falaki}. Local field potentials and behavioral data were recorded from two adult male \textit{Macaca mulatta} monkeys. These macaques were trained on a custom-made reach-to-grasp behavioral task. Given instruction cues, they had to reach with the left or right hand and apply pressure to the force transducers using a vertical precision grasp. The force transducers were placed above the starting position (home plate) in front of the custom-made primate chair. The force transducers had several grips that were changed between trials; the precision grip, and the power grip. Grasping the grips required a specific pronation of the forearm, the angle of this pronation could be manipulated by adjusting the grip orientation (0\textdegree,  45\textdegree, 90\textdegree, 135\textdegree).

In each trial, the monkey had to place both hands in the starting position (home plate). After a variable delay, the left and/or right LEDs were turned on to inform which hand(s) to use. Then (between 100-1000ms), the LEDs turned off, signaling the GO cue. With the GO signal, the monkey reaches for the transducer and applies force up to a threshold. When the threshold is reached, an auditory signal is played and a drop of juice was rewarded to the monkey through a straw fixed in front of the mouth. Lastly, the monkey had to release the grip and place the hand (or hands) back on the home plate to begin the next trial.

%The \textbf{second dataset} used was from the Kaggle Seizure Prediction Challenge of the American Epilepsy Society \cite{seizure-prediction}. Intracranial EEG (iEEG) was recorded from five dogs with naturally occurring epilepsy. The iEEG was sampled from 16 electrodes at 400Hz, and then the voltage diferences were referenced to the group average. The recordings were performed using an ambulatory monitoring system that spanned multiple months to a year in some dogs, recording up to a hundred seizures in some of the subjects.

%Additionally, this Kaggle dataset includes a pair of human patients with epilepsy with iEEG monitoring to prevent future seizures. These two recordings have a varying number of electrodes and are sampled at 5000Hz, with recorded voltages referenced to a ground electrode outside the brain scalp. 

\subsection{Dataset Analyses}

Given the previous datasets, we preprocessed them using the NeuroClean pipeline and assessed the results through the Machine Learning pipeline. In particular, we used the ROC area under the curve and the accuracy of the models to assess the improvement of the pipeline in each case study. In addition, we used a particular analysis for each case study; for the first dataset, we extracted the features needed to compute the relevant electrodes to classify the phases and segmented by bands %[TODO the other datasets]. 
Lastly, an evolution of accuracy by step of the pipeline is shown.

\paragraph{Case Study 1: Motor Classification} 

The main objective of this model is to classify the data into three different classes related to the reach-to-grasp behavioral task. This data was segmented into Pre-Reach, Reach, and Grasp phases centered and cut-off at 500 timepoints per trial, per phase. The features used for the classification was the mean spectral amplitude of the different channels, defined as the expected value of the absolute value of the voltage differences of the time series per channel.

\begin{figure}[ht]
\begin{center}
\setlength{\unitlength}{1pt}
\footnotesize
\includegraphics[width=1.0\textwidth]{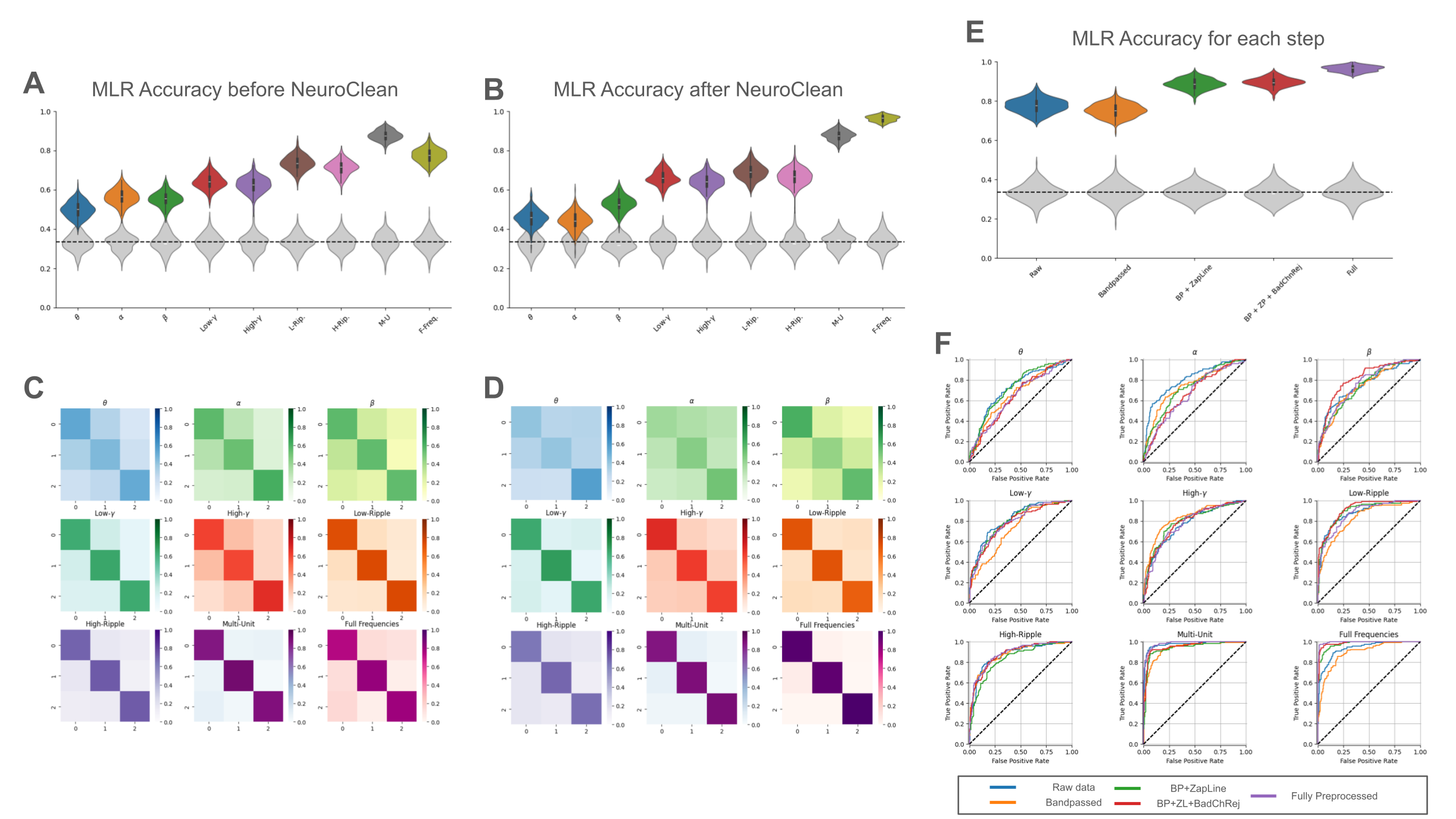}
\caption{\textit{Motor-state classification performance results. Three states were defined and classified using a multinomial logistic regressor model. \textbf{A:} A distribution of training accuracies across one hundred train-test splits for the multinomial logistic regressor (MLR) classifier applied to the data before any preprocessing was performed for all frequency bands and the full dataset. \textbf{B:} The same as \textbf{A} but for to the data after the NeuroClean pipeline was applied. \textbf{C:} Normalized confusion matrices for each frequency band for the raw unprocessed data. \textbf{D:} Same as \textbf{C} but for the data after the NeuroClean pipeline was applied. \textbf{E:} Overall distribution of accuracies after each step of the NeuroClean pipeline (Raw unprocessed data; Bandpassed only; Bandpassed and Zaplined; Bandpassed, ZapLine and with Bad Channel Rejection; fully preprocessed). \textbf{F:} The receiver operating characteristic curves for the MLR classifier for all frequency bands and computed as the micro-averaged One-Vs-Rest per step of the NeuroClean pipeline.}}
\end{center}
\label{fig:results_motor}
\end{figure}

The data was also segmented into eight different frequency bands ($\theta$ at 4Hz-7Hz, $\alpha$ at 8Hz-15Hz, $\beta$ at 15Hz-30Hz, low-$\gamma$ at 30Hz-70Hz, high-$\gamma$ at 70Hz-100Hz, low-ripple at 100Hz-150Hz, high-ripple at 150Hz-200Hz, and multi-unit at 200Hz-500Hz similar to \cite{depass}) and trained the MLR and 1NN models with all the bands and the complete frequencies data. We obtained the best accuracies in the complete frequencies with the complete pipeline at $\sim$0.97 for MLR and $\sim$0.86 for 1NN, and the second highest in the multi-unit band with $\sim$0.89 for MLR and $\sim$0.78 for 1NN. However, the multi-unit band rose from $\sim$0.81 for MLR and $\sim$0.55 for 1NN for the raw data, and from $\sim$0.78 for MLR and $\sim$0.56 for 1NN for the full frequencies for the raw data. We found that the lower frequencies than the multi-unit band were not very useful when classifying the phases (Pre-Reach, Reach, Grasp) of this motor task. Furthermore, we computed the microaveraged One-Vs-Rest ROC-AUC curves for each frequency band and data step of the pipeline, and the ROC-AUC curves show an increase through each step of the pipeline. Similarly to the accuracy, the area under the curve was maximum in the full frequencies data with >0.99 after the pipeline and 0.91 before the pipeline, see \ref{fig:results_motor} \textbf{\textit{F}} bottom right ROC-AUC curves. For the multi-unit frequency band, the ROC-AUC started directly at 0.96 and increased up to 0.98 after preprocessing. The highest precision was obtained in the Grasp state for the full frequency data with NeuroClean applied at almost perfect classification (>0.999), see \ref{fig:results_motor} \textbf{\textit{D}} bottom right square on the full band.

On the other hand, we computed the similarity to the $1/f$ curve with the resulting power spectral densities after each of the pipeline steps. The highest score was yielded by the fully preprocessed data with 0.34, compared to the 0.17 score obtained by the raw data. In the Cluster-MARA step of the pipeline, 4 clusters were found and 13 ICA components were rejected based on the probabilities of being an outlier determined by the DBSCAN algorithm. For this specific study, the median number of channels rejected was 27 out of 256 channels, and the median number of ICA components rejected was 15.

\section{Discussion}

The conditioning of EEG and LFP recording datasets yields several challenges.
In particular, these data contain a high degree of physiological and nonphysiological noise. Previous work has created several pipelines and methods to remove particular noise and artifacts from the neural data (i.e ADJUST, HAPPE, ZapLine). These pipelines usually adjust to a particular scenario and with varying degree of automation, processing the data in a fashion agnostic to the context 
of study and of the tasks that will be performed with them. Furthermore, they 
most often require specific supervised models or electrode configurations that
may hinder the generalization of the pipelines.

By contrast, the NeuroClean pipeline offers a fully unsupervised and modular solution designed to generalize across diverse datasets. The key element in 
this context, is that NeuroClean tailors the cleaning process to the specific
task and context of study.
By combining traditional signal conditioning methods with automated bad channel rejection and the novel Cluster-MARA algorithm, NeuroClean effectively removes artifacts without sacrificing neural signal integrity. The balance is reflected in the remarkable improvement in accuracies from raw to cleaned data, and in enhanced ROC-AUC values and spectral density similarities to the $1/f$ curve.

A central methodological contribution of NeuroClean is the introduction of the unsupervised Cluster-MARA algorithm, which replaces traditional supervised space-location-dependent component artifact rejection with a DBSCAN clustering of MARA features within the ICA component space. This design removes the dependence on external spatial features or training datasets, enhancing the geneneralization 
for those datasets that do not have this metadata available. The performance of DBSCAN is sensitive to parameter selection (such as neighborhood radius and minimum sample count) and the assumption that artifacts manifest as statistical outliers may not always be true. In scenarios involving structured and highly correlated noise, more adaptive cluster rejection strategies may be necessary. Likewise, the bad channel rejection step relies on fixed thresholds that may be subject to change depending on the dataset. However, this parameterization is fast as both the Cluster-MARA and Bad Channel Rejection steps are computationally fast. Thus, a quick sanity check can be performed on the data with different parameters. Another constraint concerns the validation framework in this study. 
NeuroClean was evaluated primarily through classification accuracy and machine
learning performance, which, although robust indicators of signal quality, do 
not fully capture the degree of signal preservation of brain activity temporal
dynamics. However, with this approach, NeuroClean ensures that task-relevant 
data is preserved during preprocessing.

In summary, NeuroClean provides a reproducible, unsupervised, and extensible preprocessing pipeline that improves the quality and interpretability of the
data and of subsequent analyses while minimizing and eliminating human bias 
and labor requirement. Its data-driven
architecture and demonstrated empirical performance make it a promising foundation for a standardized neural preprocessing pipeline. Continued refinement and broader validation with more datasets will further build confidence in the NeuroClean 
as a versatile tool for scalable neuro-physiological research.

\printbibliography

%\bibliographystyle{alpha} 
%\bibliography{samplebib}
%inline the .bbl file directly for mailing to authors.

%\begin{thebibliography}{Com79}

%\bibitem[Com79]{Comer-btree}
%D.~Comer.
%\newblock The ubiquitous b-tree.
%\newblock {\em Computing Surveys}, 11(2):121--137, June 1979.

%\bibitem[Knu73]{Knuth-vol3}
%D.~E. Knuth.
%\newblock {\em The Art of Computer Programming -- Volume 3 / Sorting and
%  Searching}.
%\newblock Addison-Wesley, 1973.

%\end{thebibliography}

\end{document}